\documentclass[sigconf]{acmart}

\usepackage{algorithmic}
\usepackage{graphicx}
\usepackage{subcaption} % Preferred over subfigure
\usepackage{textcomp}
\usepackage{xcolor}
\usepackage{makecell}
\usepackage{multirow}
\usepackage{enumitem}
\usepackage{booktabs}
\usepackage{tcolorbox}
\usepackage{verbatim}
\usepackage{listings}
\usepackage{float}
\usepackage{varwidth}

\renewcommand\footnotetextcopyrightpermission[1]{} % removes footnote with conference information in first column

\AtBeginDocument{%
  \providecommand\BibTeX{{%
    \normalfont B\kern-0.5em{\scshape i\kern-0.25em b}\kern-0.8em\TeX}}}

\setcopyright{acmlicensed}
\copyrightyear{2024}
\acmYear{2024}
\acmDOI{XXXXXXX.XXXXXXX}

\author{Fanwei Zhu}
\authornotemark[1]
\email{zhufw@hzcu.edu.cn}
\affiliation{%
  \institution{Hangzhou City University}
  \city{Hangzhou}
  \country{China}
}

\author{Jinke Yu}
\authornotemark[2]
\email{yujinke.yjk@alibaba-inc.com}
\affiliation{%
  \institution{Alibaba Group}
  \city{Hangzhou}
  \country{China}
}

\author{Zulong Chen}
\authornotemark[2]
\email{zulong.czl@alibaba-inc.com}
\affiliation{%
  \institution{Alibaba Group}
  \city{Hangzhou}
  \country{China}
}

\author{Ying Zhou}
\authornotemark[3]
\email{zhouying@zhejianglab.org}
\affiliation{%
  \institution{Zhejiang Lab}
  \city{Hangzhou}
  \country{China}
}

\author{Junhao Ji}
\authornotemark[2]
\email{jijunhao.jjh@alibaba-inc.com}
\affiliation{%
  \institution{Alibaba Group}
  \city{Hangzhou}
  \country{China}
}

\author{Zhibo Yang}
\authornotemark[4]
\email{yangzhibo450@gmail.com}
\affiliation{%
  \institution{Alibaba Cloud}
  \city{Hangzhou}
  \country{China}
}

\author{Yuxue Zhang}
\authornotemark[2]
\email{yuxue.zyx@alibaba-inc.com}
\affiliation{%
  \institution{Alibaba Group}
  \city{Hangzhou}
  \country{China}
}

\author{Haoyuan Hu}
\authornotemark[5]
\email{haoyuan.huhy@antgroup.com}
\affiliation{%
  \institution{Ant Group}
  \city{Hangzhou}
  \country{China}
}

\author{Zhenghao Liu}
\authornotemark[6]
\email{liuzhenghao@mail.neu.edu.cn}
\affiliation{%
  \institution{Northeastern University}
  \city{Shenyang}
  \country{China}
}

\acmISBN{978-1-4503-XXXX-X/18/06}

\begin{document}
% \title{Automating Resume Analysis with Layout-Aware Parsing and LLM-Driven Semantic Understanding}
%\title{Leveraging In-context Features and Layout-Aware Enhancement for Resume Information Extraction with Large Language Models }
\title{Layout-Aware Parsing Meets Efficient LLMs: A Unified, Scalable Framework for Resume Information Extraction and Evaluation}

%\title{Layout-Aware Parsing Meets Lightweight LLMs: A High-Accuracy, Low-Latency Resume Extraction and Evaluation Framework at Industrial Scale} 
\begin{abstract}
Automated resume information extraction is critical for scaling talent acquisition, yet real-world deployment faces three major challenges: the extreme heterogeneity of resume layouts and content, the high cost and latency of large language models (LLMs), and the lack of standardized datasets and evaluation tools. In this work, we present a layout-aware and efficiency-optimized auto-extraction and evaluation framework to addresses all three challenges. Our system combines a fine-tuned layout parser to normalize diverse document formats, an inference-efficient LLM extractor based on parallel prompting and instruction tuning, and a robust two-stage automated evaluation framework supported by new benchmark datasets.
Extensive experiments show that our framework significantly outperforms strong baselines in both accuracy and efficiency. In particular, we demonstrate that a fine-tuned compact 0.6B LLM achieves top-tier accuracy yet significantly reduces inference latency and computational cost. The system is fully deployed in Alibaba’s intelligent HR platform, supporting real-time applications across business units.

\end{abstract}
%a practical and scalable solution for industrial-strength resume information extraction.


\maketitle
\allowdisplaybreaks

\newcommand{\eg}{\textit{e.g.}}
\newcommand{\xeg}{\textit{E.g.}}
\newcommand{\ie}{\textit{i.e.}}
\newcommand{\xie}{\textit{I.e.}}
\newcommand{\etc}{\textit{etc.}}
\newcommand{\etal}{\textit{et al.}}
\newcommand{\wrt}{w.r.t. }
\newcommand{\mb}[1]{\mathbf{#1}}
\newcommand{\mc}[1]{\mathcal{#1}}
\newcommand{\figref}[1]{Fig.~\ref{#1}}
\newcommand{\tableref}[1]{Table~\ref{#1}}
\newcommand{\sectref}[1]{Sect.~\ref{#1}}
\newcommand{\B}[2]{\mathcal{B}_{#1}^{#2}} % behavior sequences of user u (#1) before time t (#2)
\newcommand{\xs}[2]{\mathcal{S}_{#1}^{#2}} % stage sequences of user u (#1) before time t (#2)
\newcommand{\s}[2]{s_{#1}^{#2}} % stage of user u (#1) at time t (#2)
\newcommand{\st}[2]{\mathbf{\tau}_{#1}^{#2}} % stage transition probability of user u (#1) at time t (#2)
\newcommand{\x}[1]{\mathbf{x}_{#1}} % basic features of user u (#1) 
\newcommand{\uv}[2]{\mathbf{v}_{#1}^{#2}} % embedding vec. of user u (#1) at time t (#2)
\newcommand{\iv}[1]{\mathbf{v}_{#1}}  %embedding vec. of item i (#1)
\newcommand{\fs}{g}   %mapping fun. for stage transition
\newcommand{\samp}{\mathbb{Y}}
\newcommand{\stitle}[1]{\vspace{2mm} \noindent {\bf #1}}
\newcommand{\sstitle}[1]{\vspace{1.5mm} \noindent {\emph{#1}}}
\newcommand{\ours}{\textbf{$M3$Net}}

%vecotrs
\newcommand{\priceV}{\vec{z}}
\newcommand{\qualityV}{\vec{q}}
\newcommand{\incomV}{\vec{i}}
\newcommand{\excomV}{\vec{e}}
\newcommand{\lpopV}{\vec{l}}
\newcommand{\spopV}{\vec{u}}
\newcommand{\comV}{\vec{c}}
\newcommand{\popV}{\vec{p}}
%\newtheorem{problem}{Problem}[]

%used vectors: h, v, x, w, y, m
\section{Introduction}
The ability to efficiently and accurately screen resumes has become a critical part of the recruitment process in modern enterprises. 
However, manual review is slow, costly, and prone to error, making it impractical for industrial use. While automated resume analysis offers a solution, existing methods often struggle to balance accuracy, latency, and computational cost.

Early resume parsing systems, built on handcrafted rules or traditional statistical models~\cite{yu2005resume,chen2016information}, offer fast processing but fail to generalize across the vast diversity of linguistic styles and visual layouts found in real-world resumes. Conversely, modern LLMs~\cite{gemini2.5_2024,claude4_2024,deepseek2024} provide the deep semantic understanding needed for robust extraction, but their high inference latency and computational expense are often prohibitive for real-time, large-scale deployment. An effective industrial system must therefore bridge this gap, delivering state-of-the-art accuracy without compromising on production efficiency and cost.

\stitle{Challenges.}
Specifically, building a practical resume analysis system at industrial scale requires addressing three key challenges
\begin{itemize}[leftmargin=*]
    \item \textit{Layout and content heterogeneity}: Resumes submitted by candidates are highly diverse in both structure and content. Many contain important information embedded in images or use complex, multi-column formats that disrupt standard reading order, making consistent parsing difficult.

    \item \textit{High cost and latency of LLMs}: While LLMs offer strong extraction capabilities, directly applying them to raw text leads to high latency and token usage, which is costly and unsuitable for real-time, large-scale applications.
    \item \textit{Lack of data and evaluation tools}: Due to privacy concerns, high-quality annotated resume datasets are rare. Evaluating extraction quality, especially for list-style entities like work experience, is hard to do manually at scale, calling for automated and reliable evaluation frameworks.
\end{itemize}

In this work, we present a practical, layout-aware, and efficiency-optimized framework for automatic resume extraction and evaluation. Our system addresses the above challenges through the following key components:
\textit{First,} we introduce a unified layout parsing model that fuses PDF metadata with OCR content and employs a fine-tuned resume layout parser to reconstruct a semantically coherent reading order from diverse, often multi-column resume layouts. \textit{Second,} to enable efficient LLM-extraction, we adopt a task decomposition strategy with index-based pointer outputs, reducing both token usage and response time. On top of this, we fine-tune a compact 0.6B model using instruction supervision, enabling high accuracy at low cost.
\textit{Third,} we develop a two-stage evaluation framework using the Hungarian algorithm for entity alignment and multi-strategy field matching. This enables robust, fine-grained assessment without human involvement.

Extensive experiments on a synthetic dataset with diverse layouts and a complex real-world resume dataset demonstrate that our system consistently outperforms state-of-the-art baselines in accuracy and efficiency. Notably, our fine-tuned Qwen3-0.6B-SFT model surpasses the accuracy of top-tier models like Claude-4 while offering 3–4$\times$ faster inference. The complete system is deployed within Alibaba Group’s intelligent HR platform, where it supports real-time parsing with high throughput across multiple business units.

\stitle{Contributions.}
In summary, our key contributions are as follows:

\begin{itemize}[leftmargin=*]
\item We propose a unified, layout-aware resume parsing framework that robustly handles layout and content heterogeneity.
\item We design an inference-efficient LLM extraction strategy and fine-tune a compact model to achieve competitive accuracy at low latency and cost.
\item We introduce a robust two-stage automated evaluation protocol for field-level performance measurement.
\item We open-source the full pipeline, along with the datasets to promote future research and practical adoption.\footnote{The code and dataset will be released at Alibaba Open Source: \url{https://github.com/ALIBABA} upon completion of the internal approval process.}
\end{itemize}

\section{Related work}
\label{sec:related}

\stitle{Rich document understanding.}
Resume information extraction is closely related to the broader field of visually-rich document understanding. Traditional NLP models, which process text as a 1D sequence, often fail on semi-structured documents like resumes, invoices, and forms.
Recent advances in this area have been driven by multimodal models that jointly captured textual content and layout structure.
 Xu \etal ~\cite{xu2020layoutlm} introduced a pre-training model LayoutLM that integrates text content with bounding box information, overcoming limitations of purely text-based NLP models on various document understanding tasks. 
LayoutLMv2~\cite{xu2020layoutlmv2} further enhanced this approach by incorporating visual features from raw document images alongside text and layout signals. LayoutLMv3~\cite{huang2022layoutlmv3} unified token and patch representations in a single Transformer, demonstrating improved multimodal reasoning.

%Follow-up works extends LayoutLM by incorporating text, layout, and visual features from document images in a multimodal model~\cite{xu2020layoutlmv2} and processes text tokens and raw image patches in a unified Transformer, enabling more effective multimodal learning~\cite{huang2022layoutlmv3}.  
Recent efforts have also explored more efficient alternatives. Wang \etal~\cite{wang2023docllm} proposed to integrate spatial structure into a language model using disentangled attention, avoiding the use of a heavy vision encoder. Zhang \etal~\cite{zhang2025sail} focused on training-free prompting strategies for LLMs, using entity, layout, and document similarities to construct more effective in-context examples.
%While these approaches target general document types, our work focuses specifically on resumes, addressing their unique challenges as discussed in Sect.~\ref{sec:intro}.

%Zhang \etal~\cite{zhang2025sail} develop a training-free framework for document information extraction that builds in-context prompts using entity-level, layout, and document-level similarities, enabling LLMs to better understand both text and structure.

\stitle{Automatic Resume Analysis.}
%Traditional resume parsing often relied on rule-based systems or template matching, which proved brittle when faced with unseen formats. Subsequent approaches leveraged machine learning techniques such as Conditional Random Fields (CRFs) and Support Vector Machines (SVMs) for sequence labeling tasks like named entity recognition (NER) and section identification. While more robust than rule-based methods, these models often required extensive feature engineering and struggled with complex, non-linear layouts.
%Beyond direct extraction, research has also explored resume classification, summarization, semantic analysis for candidate-job matching, and the development of broader AI-driven resume analysis and recommendation systems.
The automated analysis of resumes has a rich history, with research evolving from traditional rule-based systems to modern neural models.
Early approaches framed it as a Named Entity Recognition (NER) problem and solved it with hierarchical extraction pipelines. For example, Yu~\etal~\cite{yu2005resume} proposed a two-pass cascaded hybrid model that first used a Hidden Markov Model (HMM) to segment the resume into general sections (\eg, Education, Experience) and then applied specialized models (HMM or SVM) within each block.
Chen~\etal~\cite{chen2016information} advanced the cascaded approach by explicitly incorporating PDF-specific layout features (\eg, font size, coordinates) into both their SVM-based block classifier and their CRF-based field extractor. 
More recently, Zu \etal~\cite{zu2019resume} modernized this pipeline by replacing feature-engineered models with neural networks. Their system first employs neural classifiers for a sophisticated line-by-line segmentation of the resume into text blocks, which are then processed by a BiLSTM-CNN-CRF model for final entity extraction.
Beyond direct extraction, research has also explored more complex and application-oriented tasks. For instance, Pawar \etal ~\cite{pawar2021extraction} moved beyond simple NER to tackle the extraction of complex, N-ary, cross-sentence relations, using a joint hierarchical neural model. Daryani \etal \cite{daryani2020automated} built an end-to-end candidate ranking system using an NLP parser followed by IR-based resume-to-job matching. Ali \etal~\cite{ali2022resume} focused on document-level classification to sort resumes into job categories using SVMs.

%Our framework inherits this successful cascaded philosophy but modernizes it with state-of-the-art components: we replace heuristic or simple classifier-based segmentation with a more robust, fine-tuned layout parser, and we substitute traditional extractors with a powerful, efficiency-optimized LLM.

While prior methods have advanced the field, our work uniquely targets the practical challenges of deploying resume extraction systems in real-world hiring workflows. We explicitly address the layout and content heterogeneity of resumes, the high latency and cost of LLM inference at scale, and the inefficiency of manual evaluation—enabling accurate, scalable, and production-ready resume analysis for industrial use.

\section{Approach}
\label{sec:approach}
\begin{figure*}
    \centering
\includegraphics[width=0.95\linewidth]{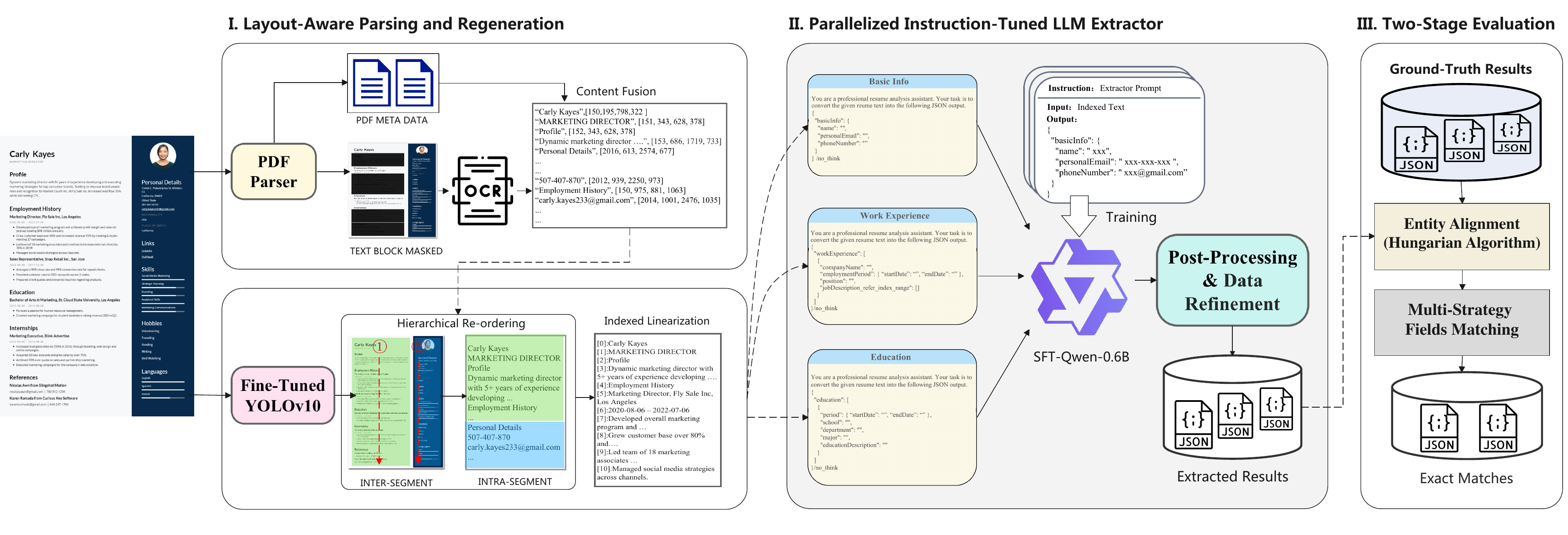}
    \caption{Overview of the layout-aware, LLM-powered resume extraction and evaluation pipeline.}
    \label{fig:archi}
\end{figure*}

Our layout-aware automatic resume information extraction system follows a three-stage architecture as illustrated in Figure~\ref{fig:archi}. First, a \textit{Layout-Aware Parsing and Regeneration} stage performs hybrid content fusion (integrating PDF metadata and OCR) and uses a fine-tuned layout regenerator to re-order text from complex, multi-column layouts into a single, indexed sequence. Second, a \textit{Parallelized, Instruction-Tuned LLM Extractor} processes this normalized text, using parallelized task decomposition and an index-based pointer mechanism with our Qwen-0.6B-SFT model to balance high accuracy with low-latency inference. Finally, a \textit{Two-Stage Automated Evaluation framework} ensures objective assessment by first using the Hungarian algorithm for robust entity alignment and then applying a multi-strategy matching logic for a fine-grained, field-level comparison.

\subsection{Layout-Aware Parsing and Regeneration}
\subsubsection{PDF Parser: Text Extraction \& Fusion}
Resume files submitted by candidates vary widely in format such as PDF, Word, \etc. To support consistent downstream processing, we first convert all files to PDF for unified processing. However, PDF resumes still include challenges such as embedded images, non-standard fonts, and custom encodings that hinder reliable text extraction. To address this, we adopt a hybrid content extraction strategy that combines metadata-based parsing with OCR.

\stitle{Hybrid PDF Content Extraction.} We extract available metadata from the PDF, which typically includes structured text content and associated positional metadata (\eg, bounding boxes) for each text object. In parallel, we render each PDF page into an image and identify candidate image regions by masking out known text regions using metadata bounding boxes. Remaining unmasked areas are treated as image regions and processed with a state-of-the-art OCR engine. The OCR results are then spatially aligned back to their corresponding coordinates within the document.

\stitle{Content Fusion.} The final step in this stage is to fuse the textual content obtained from the metadata-based PDF parsing with OCR-derived image region text. 
 The result is a single, layout-aware set of content primitives. Each primitive is a tuple containing a text string and its corresponding bounding box coordinates (text, $x_{min}$, $y_{min}$, $x_{max}$, $y_{max}$). This fusion ensures that all textual information from the resume is captured, creating a complete, but structurally unordered collection of resume content that serves as the input for our layout unification stage.

\subsubsection{Layout Regeneration: Fine-tuned Layout Parser \& Unification}
\label{subsec:layout}

Having obtained a complete but unordered set of text primitives, the next critical task is to reconstruct a semantically coherent reading flow. 

Our empirical analysis shows that approximately $20\%$ of resumes employ non-linear, multi-column layouts that break the standard top-to-bottom, left-to-right reading flow. This makes content-level parsing alone insufficient; layout understanding and unification are essential.
Conventional document layout analysis requires fine-grained annotations, which are costly and impractical for privacy-sensitive resume data. To address this, we propose a lightweight layout reconstruction approach that segments resumes into logically coherent blocks and reorders them hierarchically to produce a unified reading sequence.

\stitle{Layout Segments Identification.} 
We frame the problem of identifying layout regions (\eg, sidebars, main content columns) as an object detection task with a simplified objective: to partition a non-linear page into a set of smaller, linearly-structured layout segments. A segment is considered linear if its internal content can be correctly read by a simple top-to-bottom, left-to-right sort.

To achieve this, we fine-tune the YOLOv10 object detection model~\cite{wang2024yolov10} for the resume domain. We construct a resume-specific segmentation dataset of around 500 resumes, annotating only the bounding boxes of the major layout segments required to enforce linearity within each box. This lightweight annotation strategy avoids the burden of fine-grained labeling while effectively capturing internally linear blocks. 

%Fine-tuning YOLOv10 on this dataset yields a robust layout segment detector, achieving a mean Average Precision (mAP) of 0.921 (see Section~\ref{sec:expt}).

%This model can accurately decompose a complex, non-linear resume page into a set of discrete, internally linear ``big blocks."

\stitle{Hierarchical Re-ordering and Indexed Linearization.}
After segmenting each document into layout regions (also referred to as \textit{big blocks}), we impose a two-level hierarchical sorting strategy to regenerate a uniform layout with consistent reading order:
\begin{itemize}[leftmargin=*]
    \item Inter-segment sorting: The detected layout segments are sorted globally according to their visual positions (\ie, the coordinates of their top-left corners), following a standard top-to-bottom, left-to-right rule. This establishes the high-level reading flow, such as processing the main content column before a right-aligned sidebar if they start at the same vertical position.
    \item Intra-segment text block sorting: Within each layout segment, we further sort the individual text primitives (\ie, text blocks) that fall within its boundaries. This local sort also follows the top-to-bottom, left-to-right principle, arranging the lines and words into a coherent, readable sequence.
\end{itemize}

%This hierarchical approach effectively transforms the 2D layout problem into a structured, tree-like sorting task. 
These sorted blocks are concatenated to form a single, linearized text stream that conforms to human reading patterns. Noticeably, when constructing this sequence, we assign a unique, sequential index (\ie, a line number) to each line in the text block.
This indexing brings significant efficiency gain as we will discuss in the subsequent information extraction stage (Section 3.3). 

%Requesting an LLM to generate long-form text, such as a multi-sentence job description, is inefficient. By providing the LLM with the indexed text, we can instruct it to return a pointer (\ie, the start and end line numbers corresponding to that field) rather than the full text to improve the speed of the processing pipeline.

%For example, a work_description field would be extracted as a line number range, such as [15, 25], rather than the full text spanning those lines. This strategy dramatically reduces the token count of the LLM's output, thereby decreasing computational costs and improving the speed and fidelity of the overall extraction pipeline.

% This hierarchical reconstruction forms the basis for high-quality linearized input to the LLM-based information extraction stage, ensuring the text conforms to human reading expectations and semantic continuity.

%Complex Layouts: Many resumes adopt complex designs such as multi-column formats, side-by-side sections, or image-text overlays. These designs invalidate the standard top-down, left-to-right reading order assumed by linear parsers.

%\subsection{LLM-based Extractor: Parallel Prompting \& Instruction Tuning}
\subsection{Parallelized Instruction-Tuned LLM Extractor}
With layout unification completed, the document is converted into an indexed, linear text sequence. The final step is to extract structured key-value information suitable for automated recruitment systems.
To achieve higher accuracy and robustness, our framework leverages the advanced understanding and reasoning of LLMs to design an LLM-based extraction strategy with a particular focus on the efficiency optimizations.

\subsubsection{Parallelized Extraction via Task Decomposition.}
Our primary goal is to extract four categories of resume content essential for downstream HR applications: \textit{Basic Information} (\eg, name, contact details, location), \textit{Work Experience} (\eg, company, title, dates, description), and \textit{Education Background} (\eg, institution, degree, major, dates).
A na\"ive approach would be to prompt the LLM to extract all fields in a single pass. However, this method is suboptimal, as complex prompts requiring the model to identify disparate types of information simultaneously can degrade performance. Instead, we adopt a \textit{parallelized task decomposition} strategy, decomposing the task into three independent sub-tasks, one for each information category. The complete, linearized resume text is processed in three parallel threads, with each thread invoking the LLM with a highly specialized prompt tailored to its specific extraction target. For example, the prompt for basic information extraction explicitly requests name, phone, email, \etc, while the work experience prompt specifies extraction of job titles, time spans, and description. 
This parallelized, decomposed approach not only improves extraction accuracy but also significantly reduces end-to-end latency. Example prompts are provided in Appendix~\ref{app:prompts}.

\stitle{Index-based Pointer Mechanism.} To further address the challenges of high latency and content drift when extracting long, descriptive fields (\eg, \textit{work description}, \textit{project description}), we apply a key optimization of index-based pointer mechanism in our framework. Rather than prompting the LLM to generate the full verbatim content, a process that is slow, token-intensive, and prone to generative errors like paraphrasing or hallucination, we explicitly prompt it to return a line number range (\eg, [15, 25]) referencing the indexed text from our layout regeneration stage (Section~\ref{subsec:layout}).

This transforms the task for the LLM from a complex, open-ended generative task into a much simpler and more constrained span identification task, offering two major benefits:
\begin{itemize}[leftmargin=*]
    \item Efficiency: Returning index ranges requires only a few tokens, significantly reducing inference time and cost compared to generating full descriptions.
    \item Content Fidelity: During our post-processing stage, we use these returned indices to re-extract the exact text block from the original source document, guaranteeing 100\% content fidelity. 
\end{itemize}

\subsubsection{Model Selection and Fine-Tuning.}
Although large language models such as GPT-4o and Qwen-max exhibit strong generalization and semantic capabilities, their high latency and inference cost are often prohibitive for real-time, large-scale enterprise applications. In contrast, compact models like Qwen-0.6B offer significantly faster inference (approximately 150 tokens per second), enabling them to process a typical resume (300–400 tokens) in 1–2 seconds per extraction sub-task. However, the lightweight Qwen-0.6B lacks sufficient performance for complex extraction tasks due to limited capacity. 

To address this, we fine-tune Qwen-0.6B on a carefully constructed supervised dataset tailored for resume parsing. The supervised fine-tuning (SFT) dataset consists of {15,500 resumes} and {59,500 instruction-based training samples}. Each training sample is represented as a triplet of {(instruction, input, output)}, where the {instruction} specifies the extraction task (\eg, extracting work experience), the {input} is the indexed resume text, and the {output} is corresponding human-verified JSON-format label. The dataset combines both synthetic and real-world resumes covering diverse formats and content styles. Further construction details are provided in Appendix~\ref{app:sft_data}. 
After fine-tuning, the adapted Qwen-0.6B model achieves strong extraction accuracy comparable to much larger LLMs, while maintaining high inference efficiency, making it practical and scalable for real-time resume information extraction in industrial deployment.

\stitle{Output Formatting.}
To decide the suitable output format, we conducted experiments comparing several formats and adopt JSON as the output format. While alternatives like YAML or Markdown reduce token length, JSON yields the most stable results, likely due to Qwen’s fine-tuning on JSON-style data. Instead of using JSON-compatible decoding modes (\eg, automaton decoder), which often compromise the model's parsing capabilities, we adopt a prompt-based strategy that naturally guides the model to produce valid JSON and apply lightweight string-based extraction (\eg, text.find("\{") and text.rfind("\}")) to retrieve valid JSON blocks. 
%This strategy preserves the model's full reasoning capabilities while ensuring reliable, structured output.

\subsubsection{Post-Processing and Data Refinement}
As the raw model outputs often contain formatting inconsistencies or hallucinated content, we further implement a multi-stage post-processing and data refinement pipeline to enhance the data fidelity.
%To ensure the final data meets the high standards of accuracy and reliability required for our industrial application, we that transforms the model's raw JSON output into a clean, verified, and structured dataset.
The process consists of four key stages:
\begin{itemize}[leftmargin=*]
    \item Grounded content re-extraction: To mitigate content drift in long-form fields (\eg, \textit{job} or \textit{project description}s), we use the line indices returned by the LLM as pointers to re-extract all descriptive text directly from the original source document, eliminating any content drift introduced by the model.
    \item Domain-specific normalization: We apply a set of domain-specific cleaning and normalization rules to standardize volatile fields, such as normalizing diverse date formats and removing suffix noise in organization names.
    \item Context-aware de-duplication: We identify and filter redundant entries, such as a project that is also mentioned within a work experience, by comparing the source text spans (\ie, line number ranges) of all extracted entities.
    \item Source text verification: Finally, we perform a verification step for every extracted record, discarding any entity whose key identifying fields (\eg, \textit{company name} and \textit{job title}) cannot be found in the original document text, thereby pruning model hallucinations.
\end{itemize}
\vspace{-1.5mm}
%A detailed description of each stage in this pipeline is provided in Appendix~\ref{app:postprocess}.

Through this rigorous four-stage pipeline, we transform raw LLM outputs into clean, trustworthy, and application-ready structured data that faithfully reflects the content of the original document.

\begin{comment}
\sstitle{Output formatting.}
To decide the suitable output format, we conducted experiments comparing several formats, including JSON, XML, YAML, and Markdown. While YAML and Markdown yielded a 15-20\% reduction in output tokens, we observed that JSON provided the most stable and reliable parsing results. This is likely due to that the Qwen-2.5 model series has been specifically fine-tuned on JSON-formatted data. 

Furthermore, we found that using JSON-compatible decoding modes (\eg, automaton decoder), sometimes compromised the model's parsing capabilities. To address this, we enforce output formatting via natural prompting and use simple post-processing to extract JSON block from the model's output using string delimiters (\eg, text.find("\{") and text.rfind("\}")). 
This strategy preserves the model's full reasoning capabilities while ensuring reliable, structured output.
\end{comment}

\subsection{Two-Stage Automatic Evaluation}
\label{sec:autoEval}

After extracting the key information, another critical task is to evaluate the accuracy of extraction. Manual evaluation is impractical for large-scale system development as it is prohibitively time-consuming, expensive, and often subject to human inconsistency. To enable efficient comparison and provide objective, reliable results, an automated evaluation methodology is essential.

For clarity, we define two key terms in resume information extraction:
\begin{itemize}[leftmargin=*]
    \item An \textit{Entity} refers to a complete block of information. For example, a single "Work Experience" or one "Education History" entry is considered one entity. %A resume typically contains a list of such entities.
    \item A \textit{Field} refers to a specific attribute within an entity. For a "Work Experience" entity, its fields would be company name, position, start date, end date, and description.
\end{itemize}

Our primary evaluation challenge is to accurately compare the list of entities, along with their fields predicted by our model against a ground truth list. However, designing such an automatic evaluation model is non-trivial. A naive approach, such as a one-to-one comparison of entities based on their sequential order often fails due to the following challenges:
\begin{itemize}[leftmargin=*]
    \item Quantity mismatch: The number of predicted entities may differ from ground-truth, leading to either missed extractions or spurious extractions.
    \item Order mismatch: Even if both lists contain the same entities, the model may extract them in an order that differs from the ground-truth, causing a naive index-based comparison to fail.
    \item Partial match: An extracted entity could be only a partial match as some fields may incorrectly extracted or even missing. For instance, a predicted \textit{Work Experience} entity might have the correct company name and position fields, but a different data format or incomplete description text.
\end{itemize}
To overcome the above challenges, we propose a robust, two-step automated evaluation framework that intelligently aligns entities, performs a fine-grained field-level comparison, and aggregates the results into clear, quantitative metrics.

\stitle{Entity Alignment via the Hungarian Algorithm.}
For any pair of lists to be compared, for instance, a ground-truth list of 
$M$ work experiences $G=\{g_1, \ldots, g_M\}$ and a predicted list of $N$ experiences $P=\{p_1,\ldots,p_N\}$, we construct an $M\times N$ similarity matrix $S$. Each element $S_ij$ represents the similarity between $g_i$ and $p_j$, which is computed as the average normalized string similarity of their key fields, such as company name and position. 
We then apply the Hungarian algorithm~\cite{kuhn1955hungarian} to $S$ to find one-to-one assignment that maximizes the total similarity of all matched pairs $\{(p_i, q_j)\}$. This naturally resolves the aforementioned challenges: it is impervious to order mismatch and can find the optimal partial assignment even when $M\ne N$. 
%It outputs a set of matched pairs, a set of unmatched ground-truth entities, and a set of unmatched predicted entities.

\stitle{Multi-Strategy Fields Matching.}
Once entity pairs are aligned, we proceed to a fine-grained comparison of their constituent fields for each aligned entity pair. Recognizing that a single ``exact match" rule is inadequate for diverse data types, we designed a multi-strategy comparison function that dynamically selects the most appropriate validation rule based on the field's semantic nature.

\begin{itemize}[leftmargin=*]
    \item Period Fields (\eg, dates): Normalized into (year, month) format to support flexible matching across date representations.
\item Named Entities (\eg, organizations, schools, job titles): Compared using partial substring matching to tolerate abbreviations or suffix differences.
\item Long Descriptions (\eg, project or job descriptions): Matched via edit-distance-based similarity (\eg, edit distance > 0.9) to allow for minor paraphrasing.
\item Other Fields (\eg, names, email addresses): Evaluated using normalized exact match after lowercasing and punctuation removal.
\end{itemize}
%More details about the match strategy can be found in Appendix~\ref{app:}
An extracted field is considered as correct only if it is \textit{both} successfully aligned with a ground-truth entity by the Hungarian algorithm \textit{and} subsequently passes the multi-strategy field matching criteria. To validate the reliability of this automatic evaluation framework, we conducted human verification on a subset of results. The evaluation consistently aligned with human judgment, confirming its effectiveness in producing accurate matching outcomes.

Finally, we aggregate the field-level matching outcomes across all evaluated resumes to compute quantitative performance metrics for each field. Such per-field evaluation not only aggregates to overall extraction quality but also highlights field-specific weaknesses- offering valuable insights for further model improvement.

\section{Experiments}
\label{sec:expt}
\subsection{Experimental Settings}
\subsubsection{Datasets}
As there is no public resume dataset for evaluation, we construct two distinct datasets to evaluate our model's performance. 
%for assessing generalization under practical scenarios.
\begin{itemize}[leftmargin=*]
    \item \textbf{SynthResume}. SynthResume is a synthetic corpus of 2,994 resumes with different layouts, which is constructed through a semi-automated, LLM-based generation pipeline (Figure~\ref{fig:synthResume}). This process involves curating a diverse set of resume templates and using an LLM to populate them with new, randomized yet plausible content. Each sample undergoes layout parsing and pre-labeling via Qwen-max, followed by manual correction to ensure annotation quality. Details of dataset construction is provided in Appendiex~\ref{app:dataset}.
    \item \textbf{RealResume}. RealResume is a real-world private dataset of 13,100 real-world resumes collected from Alibaba's HR system. This dataset is inherently more complex, featuring documents with challenging characteristics such as custom fonts, intricate layouts, and mixed Chinese-English content. The annotation process for this dataset followed the same methodology as for SynthResume.
\end{itemize}
\begin{comment}

\begin{table}[h]
\centering
\caption{Statistics of Datasets.}
\resizebox{\columnwidth}{!}{%
\begin{tabular}{lcc}
\toprule
\textbf{} & \textbf{SynthResume} & \textbf{RealResume} \\
\midrule
\textbf{Source} & Semi-Automated Synthetic & Real-World \\
\textbf{Size} & 2,994 resumes & 13,100 resumes\\
\textbf{Fields} & 15 fields  & 19 fields\\
\textbf{Language} & Chinese & Mixed (Eng/Chn) \\
\textbf{Layout} & Diverse Layout Templates & Complex Layouts, Custom Fonts \\
\makecell[l]{\textbf{Data Split} \\ \textbf{(Train / Test)}} & 2,500 / 494 & 13,000 / 100 \\
\bottomrule
\end{tabular}%
}
\label{tab:dataset_comparison}
\end{table}
\end{comment}
\begin{table*}[h]
\centering
\caption{Statistics of Datasets.}
\resizebox{0.9\textwidth}{!}{%
\begin{tabular}{ccccccc}
\toprule
\textbf{Dataset} & \textbf{Source} & \textbf{Size} & \textbf{Fields} & \textbf{Language} & \textbf{Layout} & \makecell{\textbf{Data Split} \textbf{(Train/Test)}}\\
\midrule
\textbf{SynthResume} & Semi-Automated Synthetic & 2,994 & 15 & Chinese & Linear \& Non-Linear Templates & 2,500 / 494\\[5pt]
\textbf{RealResume} & Real-World & 13,100 & 19 & Mixed (Eng/Chn) & More Complex, Custom Fonts & 13,000 / 100\\
\bottomrule
\end{tabular}%
}
\label{tab:dataset_comparison}
\end{table*}

\begin{comment}

\subsubsection{Methods for Comparison}
To thoroughly evaluate our proposed framework, we compare it against a wide range of baselines, including traditional NLP and LLM-based approaches. We also benchmark different LLM models in our proposed pipeline. Specifically, we organize the compared approaches into three categories: (1) \textit{Non-LLM Baselines}: A state-of-the-art industrial baseline Bello~\cite{bello2024} and a deep learning NLP pipeline PaddleNLP~\cite{paddlenlp}. (2) \textit{Na\"ive LLM Baseline}: A straightforward approach where the OCR-extracted text is fed directly to an LLM (Claude-4~\cite{claude4_2024}) without our layout-aware pipeline. (3) \textit{Our Pipeline (Zero-Shot LLM)}: Our full layout-aware pipeline using various off-the-shelf LLMs (Claude-4~\cite{claude4_2024}, Gemini-2.5flash~\cite{gemini2.5_2024}, GPT-4o~\cite{openai2024gpt4technicalreport}, Deepseek-v3~\cite{deepseek2024}, Qwen-max~\cite{qwen2024}, Qwen3 Series~\cite{qwen3_2024}) without any task-specific fine-tuning. (4) \textit{Our Pipeline (Fine-Tuned Model)}: Our full pipeline with the fine-tuned Qwen3-0.6B-SFT model.

The detailed description of each method is provided in Appendix~\ref{app:baseline}.
\end{comment}

\subsubsection{Methods for Comparison}
To comprehensively evaluate our proposed framework, we compare it against a wide range of baseline systems, as well as benchmark different LLMs in our pipeline. We categorize the compared methods into four groups:

\begin{itemize}[leftmargin=*]
    \item \textbf{Non-LLM Baselines.} This includes a commercial resume parsing system widely used in industry {Bello}~\cite{bello2024}, and a deep learning-based information extraction pipeline built on the PaddlePaddle framework {PaddleNLP}~\cite{paddlenlp}. 

    \item \textbf{Na\"ive LLM Baseline.} This approach directly applies a large language model (Claude-4~\cite{claude4_2024}) to OCR-extracted resume text, without any layout-aware preprocessing or task decomposition.

    \item \textbf{Our Pipeline (Zero-Shot LLM).} We evaluate our full layout-aware pipeline paired with various off-the-shelf LLMs, including Claude-4~\cite{claude4_2024}, Gemini-2.5flash~\cite{gemini2.5_2024}, GPT-4o~\cite{openai2024gpt4technicalreport}, Deepseek-v3~\cite{deepseek2024}, Qwen-max~\cite{qwen2024}, and Qwen3 series models~\cite{qwen3_2024}, to examine how different models perform in a zero-shot setting within our architecture.

    \item \textbf{Our Pipeline (Fine-Tuned LLM).} This variant uses our pipeline with a supervised fine-tuned model, Qwen3-0.6B-SFT, trained on our SFT dataset. It represents our final, production-oriented system, designed to balance top-tier accuracy with the efficiency required for large-scale deployment. 
\end{itemize}

Details about these models are provided in Appendix~\ref{app:baseline}.

\subsubsection{Metrics}
%After aligning entities and comparing their constituent fields across the entire dataset, we aggregate these results to compute quantitative performance metrics.

%Our evaluation is conducted on a per-field basis, which not only aggregates overall extraction quality but also reveals which types of fields are more prone to errors.
%In our experiments, we adapt three standard evaluation metrics Precision, Recall and F1-score~\cite{deepa2025automated} to resume entity extraction task, and introduce a novel metric Accuracy to specifically evaluate our multi-strategy field matching logic. 

We evaluate our system using standard information extraction metrics: \textit{Precision}, \textit{Recall}, and \textit{F1-score}, following prior work~\cite{deepa2025automated}. Evaluation is conducted at the field level, where an extracted field is considered a correct match only if it satisfies both entity alignment and field-matching criteria.
Additionally, we introduce a novel alignment \textit{Accuracy} metric that measures the fraction of correctly extracted fields among all aligned fields. This metric helps us distinguish errors caused by the alignment algorithm from those caused by our field-matching rules.

Formal definitions of the metrics are provided in Appendix~\ref{app:metrics}.

\subsubsection{Implementation Details}
We fine-tune the Qwen-0.6B model using a full-parameter supervised fine-tuning approach, updating all trainable parameters to maximize task-specific performance.
For optimization, we employ the AdamW optimizer~\cite{loshchilov2019decoupled}. The training process is stabilized by a low initial learning rate of 5e-6, which helps prevent catastrophic forgetting. The learning rate is managed by a cosine annealing scheduler~\cite{loshchilov2017sgdr} to facilitate smooth convergence in the later stages of training.
To accommodate GPU memory constraints, we set the per-device batch size to 2. We apply gradient accumulation over 2 steps, resulting in an effective batch size of 4. This configuration provides a robust balance between training stability and computational efficiency, allowing us to effectively train the model within our hardware limitations.

\subsection{Impact of Hyper-parameters}
\begin{figure}[h]
\centering
    \includegraphics[width=\linewidth]{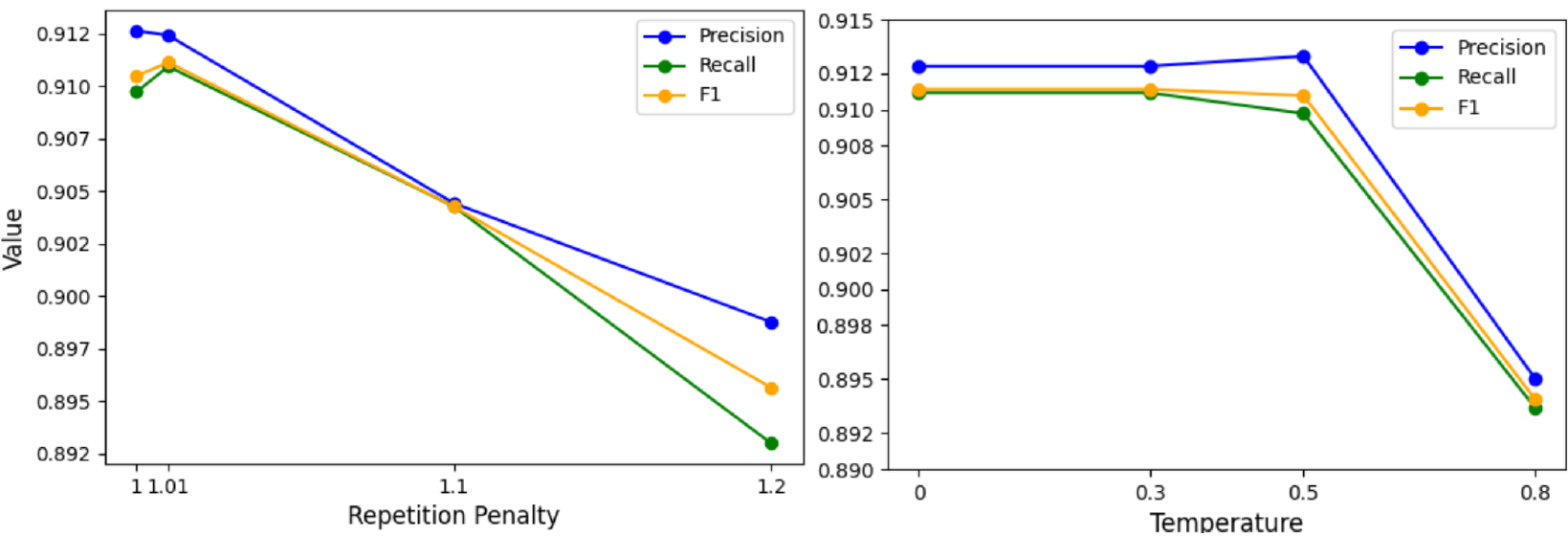}
\caption{Impact of Hyper-parameters.}
\label{fig:para}
\end{figure}

We analyze the effects of two key decoding parameters: \textit{repetition penalty} and \textit{temperature} on extraction performance. We report the results on the SynthResume dataset in in Figure~\ref{fig:para}, while similar trends observed on RealResume. 

The repetition penalty discourages the model from generating repetitive content by penalizing previously generated tokens. We vary it while keeping temperature at 0 and find that a small penalty of 1.01 yields the best F1 score. 
The temperature controls the randomness of token sampling. For this experiment, the repetition penalty is fixed at its optimal value (1.01). We find that a moderate temperature of 0.5 provides the most stable and accurate performance.
These optimal settings are adopted for all subsequent experiments.

\subsection{Performance Comparison}
We evaluate and compare the performance of various resume information extraction methods across both the SynthResume and RealResume datasets. 
The overall performance averaged across all resume fields, along with average processing time per resume, are presented in Table~\ref{tab:main_results}. We also report fine-grained accuracy comparisons across different field groups in Table~\ref{tab:finegrained}. Specifically, the \textit{Period} group includes four date-related fields: employment start/end and education start/end. The \textit{Named Entity} group includes six entity fields: company name, job title, school, major, degree, and department. The \textit{Long Text} group comprises two descriptive fields: job description and education description. From the general and fine-grained comparison, we observe the following key findings.

\begin{table*}[ht]
\centering
\caption{Comparison of Overall Model Performance on the SynthResume and RealResume Datasets.
Results are averaged across all resume fields to provide a holistic evaluation of each method. Best scores are in \textbf{bold}, second-best are {underlined}.}
\label{tab:main_results}
\resizebox{0.98\textwidth}{!}{%
\begin{tabular}{ll rrrrr rrrrr}
\toprule
& & \multicolumn{5}{c}{\textbf{SynthResume Dataset}} & \multicolumn{5}{c}{\textbf{RealResume Dataset}} \\
\cmidrule(lr){3-7} \cmidrule(lr){8-12}
\textbf{Category} & \textbf{Model} & \textbf{Acc. $\uparrow$} & \textbf{Prec. $\uparrow$} & \textbf{Recall $\uparrow$} & \textbf{F1 $\uparrow$} & \textbf{Time (s) $\downarrow$} & \textbf{Acc. $\uparrow$} & \textbf{Prec. $\uparrow$} & \textbf{Recall $\uparrow$} & \textbf{F1 $\uparrow$} & \textbf{Time (s) $\downarrow$} \\
\midrule
\multirow{2}{*}{Non-LLM  Baselines} & Bello & 0.815 & 0.787 & 0.741 & 0.762 & \underline{1.43} & {0.835} & 0.836 & 0.746 & 0.817 & \underline{1.62} \\
 & PaddleNLP & 0.576 & 0.669 & 0.474 & 0.523 & 22.0 & 0.515 & 0.584 & 0.422 & 0.492 & 20.9 \\
%& EzNLP & -- & 0.812 & 0.650 & 0.715 & 0.5 & -- & 0.911 & 0.799 & 0.849 & -- \\
\midrule
Na\"ive LLM Baseline  & Claude-4 & 0.926 &0.923 &	0.933 &	0.927  &20.54 & 0.896 & 0.896 &0.901 &	0.919 & 22.71 \\
\midrule
\multirow{8}{*}{\begin{tabular}[l]{@{}c@{}} Our Pipeline \\ (Zero-Shot LLM)\end{tabular}} 
& Claude-4 & 0.949 & 0.950 & 0.943 & 0.946 & 4.98 & \underline{0.948} & \underline{0.937} & 0.952 & \underline{0.959} & 4.62 \\
& Gemini2.5-flash & 0.949 & \underline{0.958} & 0.945 & \underline{0.951} & 11.23 & {0.947} & 0.933 & \underline{0.955} & {0.954} & 13.67 \\
& GPT-4o & \textbf{0.952} & \underline{0.958}& \textbf{0.948} & \textbf{0.952} & 5.50 & 0.944 & {0.936} & 0.950 & {0.954} & 6.26 \\
& Deepseek-v3 & \underline{0.951} & \textbf{0.959} & 0.941 & 0.950 & 8.66 & 0.939 & 0.935 & 0.936 & 0.944 & 10.58 \\
& Qwen-max & 0.950 & 0.945 & \underline{0.947} & 0.946 & 9.40 & 0.935 & 0.927 & 0.934 & 0.937 & 19.2 \\
& Qwen3-14B & 0.911 & 0.906 & 0.911 & 0.908 & 6.30 & 0.914 & 0.898 & 0.911 & 0.912 & 8.55 \\
& Qwen3-4B & 0.885 & 0.876 & 0.895 & 0.885 & 5.24 & 0.861 & 0.833 & 0.887 & 0.869 & 6.85 \\
& Qwen3-0.6B & 0.618 & 0.671 & 0.663 & 0.645 & \textbf{1.22} & 0.589 & 0.632 & 0.622 & 0.641 & \textbf{1.54} \\
\midrule
%{\begin{tabular}[l]{@{}c@{}} Our Pipeline \\ (Fine-Tuned Model)\end{tabular}} & Qwen3-0.6B-SFT & 0.931 & 0.918 & 0.917 & 0.917 & \textbf{1.22} & 0.934 & 0.920 & 0.935 & 0.940 & \textbf{1.54} \\

{\begin{tabular}[l]{@{}c@{}} Our Pipeline \\ (Fine-Tuned Model)\end{tabular}} & Qwen3-0.6B-SFT & 0.931 & 0.918 & 0.917 & 0.917 & \textbf{1.22} & \textbf{0.961} & \textbf{0.938} 	&\textbf{0.964} 	&\textbf{0.964} & \textbf{1.54}\\
\bottomrule
\end{tabular}%
}
\end{table*}
 \stitle{First, our layout-aware pipeline is critical for achieving state-of-the-art performance.} 
As shown in Table~\ref{tab:main_results}, our pipeline consistently outperforms all baselines. On the SynthResume dataset, the na\"ive LLM baseline (Claude-4) achieves an F1-score of 0.927, whereas integrating Claude-4 into our layout-aware pipeline boosts it to 0.946. The improvement is even more pronounced on the RealResume dataset, which includes more complex, mixed-language resumes- raising the F1-score from 0.919 to 0.959, a substantial 40-point gain. This significant improvement that validates the effectiveness of our framework, particularly in handling real-world resume complexity.
Moreover, our pipeline outperforms traditional non-LLM methods by a wide margin. Compared to Bello, a state-of-the-art industrial system, our fine-tuned Qwen3-0.6B-SFT model improves F1-score by over 14 points (0.817 $\rightarrow$ 0.964), while also reducing inference latency (1.62s $\rightarrow$ 1.54s), making it well-suited for real-world deployment.

\stitle{Second, our fine-tuned small model offers the optimal trade-off between accuracy and efficiency.} 
While top-tier models such as Claude-4 and Gemini-2.5 deliver strong performance, their inference latency (4–13 seconds per resume) limits scalability in high-throughput environments. Our fine-tuned Qwen-0.6B-SFT model provides the optimal balance of performance and efficiency for a production environment. As shown in Table~\ref{tab:main_results}, supervised fine-tuning boosts the base Qwen3-0.6B model’s F1-score on RealResume from 0.641 to 0.964- surpassing even Claude-4 (0.959) while reducing latency to just 1.54 seconds, achieving a 3–4$\times$ speedup.
These results validate the effectiveness of fine-tuning compact models to meet both high accuracy and low latency demands in production-scale resume processing.

%This demonstrates that SFT can effectively distill the capabilities of larger models into a compact, efficient, and production-ready asset.
\stitle{Third, the gains of our framework are most evident in complex {Long Text} fields.} Long Text fields are most challenging as they require coherent extraction of multi-sentence descriptions. As shown in Table~\ref{tab:finegrained}, on RealResume, the na\"ive LLM baseline (Claude-4) achieves only an F1-score of 0.548 in this group, but with our full pipeline, the score rises sharply to 0.854. A similar boost is seen with model fine-tuning: Qwen3-0.6B’s F1-score for \textit{Long Text} jumps from 0.136 to 0.846 after SFT. Accurately extracting long text (e.g., job descriptions and project summaries) is essential for downstream tasks such as candidate-job matching, making this improvement especially impactful in real-world hiring applications.

%This improvement stems not only from preserving semantic continuity via layout-aware parsing, but also from decomposing extraction into specialized subtasks and applying post-processing to recover faithful, well-grounded text spans. 
{Interestingly,} na\"ive LLM baseline achieves the highest F1-score on \textit{Period} fields. This suggests that LLMs may possess strong intrinsic capabilities for handling short, visually distinct, and highly regular patterns such as dates. In such cases, our layout regenerator may introduce minor segmentation noise that slightly degrades performance. This observation motivates future work on field-specific extraction strategies to further optimize performance. 
%such as applying simpler methods for regular fields while reserving our full pipeline for more complex, context-dependent ones, to further optimize performance. 

%future work in developing a dynamic, field-specific extraction strategy that could apply simpler methods for simple fields and our full, powerful pipeline for more complex, context-dependent ones.

\begin{table*}[ht]
\centering
\caption{Fine-grained Accuracy and F1 Scores on the SynthResume and RealResume datasets, grouped by field types: Period, Named Entity, and Long Text. Missing values indicate that the PaddleNLP baseline does not support extraction of Long Text fields. Full results including Precision and Recall are provided in Appendix~\ref{app:finegrained}.}

\label{tab:finegrained}
\resizebox{0.98\textwidth}{!}{
\begin{tabular}{l cc cc cc cc cc cc}
\toprule
& \multicolumn{6}{c }{\textbf{SynthResume Dataset}} 
& \multicolumn{6}{c}{\textbf{RealResume Dataset}} \\
\cmidrule(lr){2-7} \cmidrule(lr){8-13}
\multirow{2}{*}{\textbf{Model}} & \multicolumn{2}{c}{\textbf{Period}} 
& \multicolumn{2}{c}{\textbf{Named Entity}} 
& \multicolumn{2}{c}{\textbf{Long Text}} 
& \multicolumn{2}{c}{\textbf{Period}} 
& \multicolumn{2}{c}{\textbf{Named Entity}} 
& \multicolumn{2}{c}{\textbf{Long Text}} \\
\cmidrule(lr){2-3} \cmidrule(lr){4-5} \cmidrule(lr){6-7}\cmidrule(lr){8-9} \cmidrule(lr){10-11} \cmidrule(lr){12-13}
& Acc. $\uparrow$ & F1 $\uparrow$& Acc. $\uparrow$& F1 $\uparrow$& Acc. $\uparrow$& F1 $\uparrow$& Acc. $\uparrow$& F1 $\uparrow$& Acc. $\uparrow$& F1 $\uparrow$& Acc. $\uparrow$& F1 $\uparrow$\\
\midrule
\multicolumn{13}{l}{\textit{Non-LLM Baselines}} \\
\quad Bello            & 0.852 & 0.885 & 0.841 & 0.749 & 0.469 & 0.259 & 0.921 & 0.879 & 0.885 & 0.769 & 0.540 & 0.500 \\
\quad PaddleNLP        & 0.433 & 0.511 & 0.818 & 0.699 & -- & --    & 0.387 & 0.451 & 0.722 & 0.622 & --    & --    \\
\midrule
\multicolumn{13}{l}{\textit{Na\"ive LLM Baseline}} \\
\quad Claude-4   & \textbf{0.977} & \textbf{0.980} & 0.939 & {0.951} & 0.731 & 0.684 & \underline{0.979} & \textbf{0.986} & 0.937 & \textbf{0.959} & 0.582 & 0.548 \\
\midrule
\multicolumn{13}{l}{\textit{Our Pipeline (Zero-Shot LLM)}} \\
\quad Claude-4 & 0.960 & 0.973 & 0.963 & 0.956 & {0.825} & {0.801} & 0.963 & 0.972 & \textbf{0.964} & \underline{0.949} & {0.869} & {0.854} \\
\quad Gemini2.5-flash  & \underline{0.963} & \underline{0.975} & 0.959 & 0.957 & 0.835 &\textbf{ 0.831} & {0.970} & {0.978} & 0.945 & 0.931 & \textbf{0.888} & \underline{0.865} \\
\quad GPT-4o           & 0.960 & 0.974 & {0.966} & \underline{0.961} & \underline{0.840} & \underline{0.829} & 0.963 & {0.972} & {0.950} & {0.941} & {0.880} & \textbf{0.870} \\
\quad Deepseek-v3      & 0.961 & 0.971 & \underline{0.968} & 0.962 & 0.829 & 0.813 & 0.960 & 0.971 & 0.939 & 0.918 & 0.867 & 0.852 \\
\quad Qwen-max         & 0.942 & 0.945 & \textbf{0.979} & \textbf{0.970} & 0.824 & 0.811 & {0.971} & {0.978} & 0.936 & 0.915 & 0.827 & 0.820 \\
\quad Qwen3-14B        & 0.888 & 0.902 & 0.961 & 0.954 & 0.698 & 0.664 & 0.963 & 0.972 & 0.939 & 0.899 & 0.724 & 0.707 \\
\quad Qwen3-4B         & 0.893 & 0.889 & 0.955 & 0.953 & 0.513 & 0.527 & 0.901 & 0.930 & 0.921 & 0.886 & 0.579 & 0.567 \\
\quad Qwen3-0.6B       & 0.619 & 0.668 & 0.637 & 0.691 & 0.256 & 0.184 & 0.680 & 0.734 & 0.647 & 0.671 & 0.120 & 0.136 \\
\midrule
\multicolumn{13}{l}{\textit{Our Pipeline (Fine-Tuned Model)}} \\
\quad Qwen3-0.6B-SFT   & 0.897 & 0.907 & 0.951 & 0.938 & \textbf{0.858} & {0.767} & \textbf{0.981} & \underline{0.984}  & \underline{0.956} & {0.937}  & \underline{0.880} & 0.846 \\
\bottomrule
\end{tabular}
}
\end{table*}

%Robustness (performance across different layouts)

\subsection{Ablation Study}
To understand the impact of each major component in our pipeline, we conduct an ablation study with SynthResume Dataset by removing three core modules individually: (1) \textit{w/o Text Fusion}: removing PDF's metadata-based text content and relying solely on OCR for text extraction; (2) \textit{w/o Layout Generator}: Disabling the hierarchical layout reordering, a naive top-to-bottom, left-to-right sort is applied directly to the text boxes; (3) \textit{w/o Post-Processor}: Skipping the post-processing step in the LLM-based extractor, the model directly generate the full long description.
\begin{table}[H]
\centering
\caption{Ablation study on key system components.}
\label{tab:ablation}
\resizebox{\columnwidth}{!}{
\begin{tabular}{l c ccc}
\toprule
& & \multicolumn{3}{c}{\textbf{Finer-Grained Accuracy $\uparrow$}} \\
\cmidrule(lr){3-5}
\textbf{Variants} & \textbf{Overall Acc. $\uparrow$} & \textbf{Period} & \textbf{Named Entity} & \textbf{Long Text} \\
\midrule
w/o Text Fusion     & 0.907 & 0.892 & 0.945 & 0.743 \\
w/o Layout Generator           & 0.916 & 0.892 & 0.950 & 0.758 \\
w/o Post Processor  & 0.921 & 0.897 & 0.952 & 0.781 \\
\midrule
\textbf{Full system} & \textbf{0.932} & \textbf{0.897} & \textbf{0.952} & \textbf{0.858} \\
\bottomrule
\end{tabular}
}
\end{table}
As shown in Table~\ref{tab:ablation}, each component of our framework contributes meaningfully to overall performance. The full system achieves the highest overall accuracy (0.932), and removing any component leads to a clear drop, validating the effectiveness of our integrated design.
\emph{Finer-grained analysis} further shows that \textit{Long Text} fields are the most sensitive. Removing Text Fusion or Layout Generator causes over a 10-point drop in accuracy, due to OCR errors and disrupted reading order in multi-line content. Omitting the Post Processor also leads to a 7.7-point decline, highlighting the difficulty LLMs face in generating long, verbatim text accurately. Our post-processing module provides crucial robustness and traceability for such complex fields. In contrast, \textit{Named Entity} and \textit{Period} fields are more robust, but still benefit from Text Fusion, confirming that dual-modal PDF text extraction is consistently superior to OCR.

\subsection{Online Deployment}
\begin{figure}[th] 
    \centering 
    \includegraphics[width=3.3in]{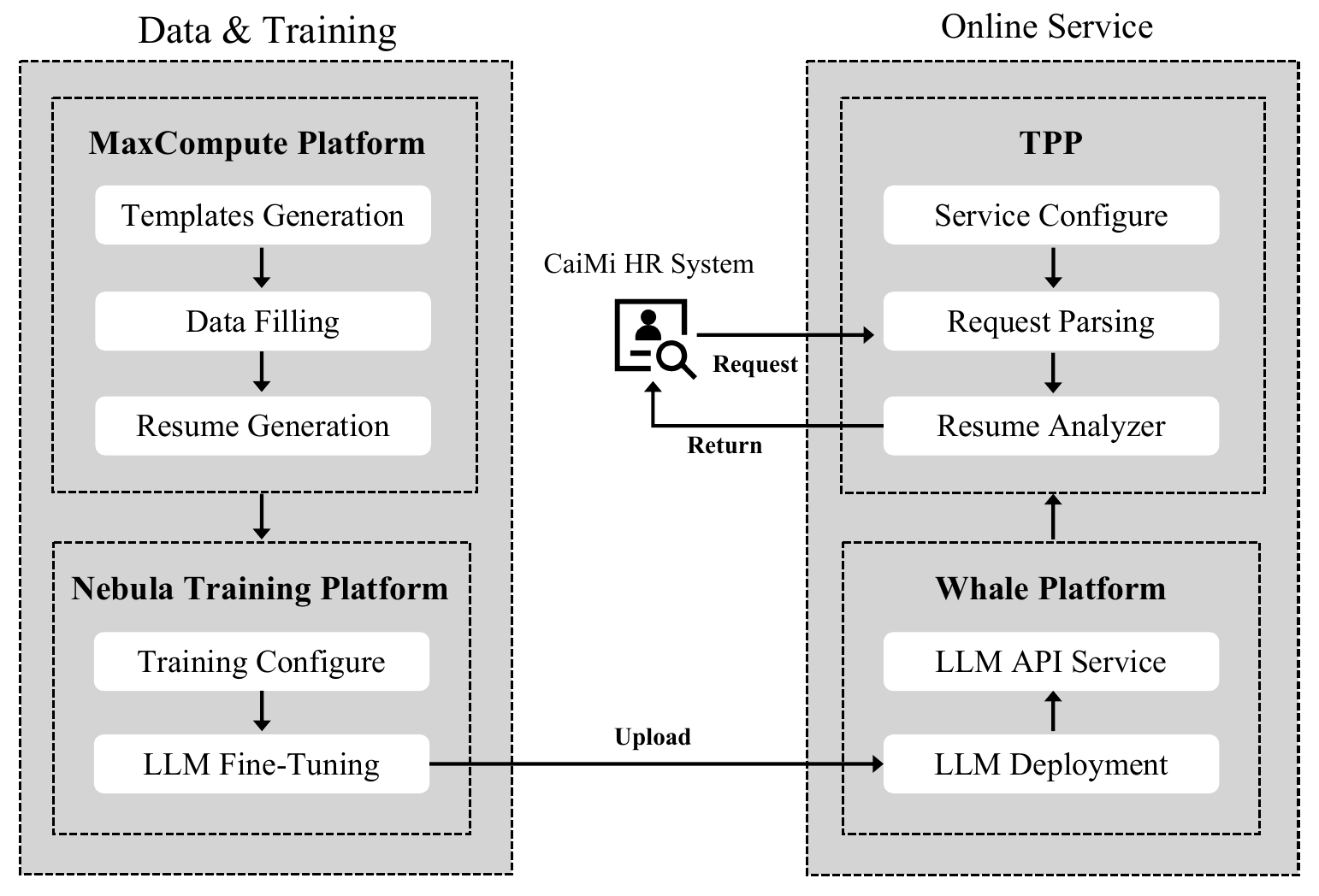} 
    \caption{System Deployment and Serving Framework.}
    \label{fig:deployment} 
\end{figure}

With the success in offline experiments, our pipeline has been deployed to support Alibaba's Intelligent HR system (CaiMi). The deployment involves both offline training and online serving, as shown in Figure~\ref{fig:deployment}.

In the \textit{offline phase}, we construct the training dataset on the MaxCompute Platform, where a diverse set of resume templates is first generated. These templates are populated using LLM-based content synthesis followed by manual correction, producing high-quality synthetic resumes. Combined with real resumes collected from Alibaba’s hiring applications, we construct a fine-tuning corpus of  labeled examples. All resume texts and labels are stored on Alibaba Cloud's Object Storage Service (OSS). Model fine-tuning is conducted on Neubla, Alibaba’s internal distributed AI training platform. We fine-tune the Qwen-0.6B model using full-parameter supervised learning on a compute node with 8× NVIDIA A800 GPUs. With Neubla’s optimized training infrastructure, the full fine-tuning process completes within 30 minutes.

At \textit{online serving}, the fine-tuned model is deployed on Whale Platform, Alibaba’s LLM-serving infrastructure. Serving orchestration is managed by TPP, Alibaba’s internal online inference engine. Upon a parsing request from the CaiMi HR system, TPP orchestrates the entire resume parsing workflow. This includes initial OCR and text fusion, layout regeneration, LLM API invocation via Whale, and finally, returning the extracted structured results back to the HR system.
The entire pipeline demonstrates strong real-time performance, achieving a throughput of 240–300 resumes per minute (\ie, 4–5 QPS), with an average response latency of 1.54 seconds per resume. 
This meets the strict latency and throughput requirements of large-scale enterprise hiring systems.

\section{Conclusion and Future Work}
In this paper, we present a layout-aware, efficiency-optimized automatic resume extraction and assessment pipeline that successfully addresses the key challenges of layout heterogeneity, LLM latency, and evaluation difficulty in industrial-scale resume information extraction. We demonstrate that by combining a robust layout-aware parsing pipeline with a lightweight fine-tuned LLM model, our approach delivers both high accuracy and low latency. The framework significantly outperforms existing baselines and has been successfully deployed in Alibaba’s intelligent HR system, serving real-time scenarios with high throughput and reliability. We also open-source the entire pipeline and contribute benchmark datasets to advance future research and practical application.
%We also propose a two-stage auto-assessment framework for fine-grained evaluation of field-level extraction quality. 
Future work will explore dynamic, field-specific extraction strategies that selectively apply different extraction models, such as applying simpler methods for regular fields while reserving our full pipeline for more complex, context-dependent ones, to further optimize performance. 
%applying simpler methods for regular fields while reserving our full pipeline for more complex, context-dependent ones,
%\clearpage
\bibliographystyle{ACM-Reference-Format}
\bibliography{custom}
\appendix
\section*{Appendix}

\section{Task-specific Prompts}
\label{app:prompts}
In the LLM-based Extractor, we adopt a parallelized task decomposition strategy, where each extraction task is handled independently using a specialized instruction prompt. This modular approach improves both extraction accuracy and efficiency in production settings. The task-specific prompts for extracting basic information, education background, and work experience are illustrated in Figures~\ref{fig:prompt_basicinfo}, \ref{fig:prompt_education}, and~\ref{fig:prompt_work_experience}, respectively.

\begin{figure*}[htbp]
\centering
\small
\begin{tcolorbox}[colback=white, colframe=black, title=Prompt for Basic Information Extraction, fonttitle=\bfseries, width=0.93\textwidth]
Extract the following information into a JSON object. If any field does not exist, output an empty string "".

\begin{verbatim}
{
  "basicInfo": {
    "name": "",            // Full name, e.g., "Zhang San"
    "personalEmail": "",   // Email address, e.g., "610730297@qq.com"
    "phoneNumber": "",     // Phone number, e.g., "13915732235".
                           // Preserve original format, including country/area code if present (e.g., "+1 (201) 706 1136")
    "age": "",             // Current age (numeric only)
    "born": "",            // Birth year and month if available, e.g., "1996-11"
    "gender": "",          // "Male" or "Female". Leave blank if not specified.
    "desiredLocation": ["city name", ...], 
                           // Explicitly mentioned preferred job location(s), e.g., ["Beijing", "Shanghai"].
                           // Only extract if clearly stated. If none, return [].
    "jobIntention": "",    // Job intention or target position, e.g., "Algorithm Engineer". Leave blank if unclear.
    "currentLocation": "", // Current city of residence. Do NOT infer from work experience or place of origin.
    "placeOfOrigin": ""    // Place of origin or hometown. Should not be confused with current location.
  }
}
\end{verbatim}

\end{tcolorbox}
\caption{Prompt for Extracting Basic Information in JSON Format.}
\label{fig:prompt_basicinfo}
\end{figure*}

\begin{figure*}[htbp]
\centering
\small
\begin{tcolorbox}[colback=white, colframe=black, title=Prompt for Education Background Extraction, fonttitle=\bfseries,width=0.93\textwidth]

Extract the following education experiences into a JSON array. If any field does not exist, output an empty string "". If no education experience is mentioned, return an empty array [].

\begin{verbatim}
{
  "education": [
    {
      "school": "",       // Full name of the educational institution, e.g., "Tsinghua University"
      "major": "",        // Major or field of study, e.g., "Computer Science"
      "degree": "",       // Degree earned, e.g., "Bachelor", "Master", "PhD"
      "startDate": "",    // Start date in "YYYY-MM" format if available, e.g., "2018-09"
      "endDate": "",      // End date in "YYYY-MM" format. If still studying, use "present"
      "location": ""      // City or region of the school, e.g., "Beijing"
    }
  ]
}
\end{verbatim}

\end{tcolorbox}
\caption{Prompt for Extracting Education Background in JSON Format.}
\label{fig:prompt_education}
\end{figure*}

\begin{figure*}[htbp]
\centering
\small
\begin{tcolorbox}[colback=white, colframe=black, title=Prompt for Work Experience Extraction, fonttitle=\bfseries,width=0.93\textwidth]

Extract the following work experiences into a JSON array. If any field does not exist, output an empty string "". If no work experience is mentioned, return an empty array [].

\begin{verbatim}
{
  "workExperience": [
    {
      "company": "",        // Company name, e.g., "Alibaba Group"
      "position": "",       // Job title or role, e.g., "Backend Engineer"
      "startDate": "",      // Start date in "YYYY-MM" format, e.g., "2020-07"
      "endDate": "",        // End date in "YYYY-MM" format; use "present" if currently employed
      "location": "",       // City or region of the job location, e.g., "Hangzhou"
      "description": ""     // Description of responsibilities and achievements; preserve original wording
    }
  ]
}
\end{verbatim}

\end{tcolorbox}
\caption{Prompt for Extracting Work Experience in JSON Format.}
\label{fig:prompt_work_experience}
\end{figure*}

%\section{Details of Fields Matching}

\section{Construction of Supervised Fine-Tuning Dataset}
\label{app:sft_data}
To adapt the Qwen-6B language model for resume information extraction, we construct a high-quality supervised fine-tuning (SFT) dataset in an instruction-based format. 
The SFT dataset consists of two parts:
\begin{itemize}[leftmargin=*]
\item \textbf{Synthetic Dataset:} We generate 2,500 synthetic resumes using a layout-diverse generation pipeline. For each resume, we create training instances for three extraction tasks (\ie, extracting basic information, work experience, and education background), yielding a total of 7,500 instruction-format samples.
\item \textbf{Real-World Dataset:} We collect 13,000 real resumes with complex formats and mixed-language content. Each resume is annotated for four extraction tasks (\ie, extracting basic information, work experience, project experience, and education background), resulting in 52,000 SFT training examples.
\end{itemize}
Each training example is represented as a triplet of {(instruction, input, output)}, where the {instruction} specifies the extraction task, the {input} is a fully indexed resume text, and the {output} is the corresponding structured JSON-formatted label derived from human-annotated ground truth. An example of instruction-based sample for basic information extraction is illustrated in Figure~\ref{fig:sample_basicinfo}.
This SFT dataset enables the model to learn fine-grained extraction behavior under explicit instructions and to generalize across diverse layout structures and resume styles.
\begin{figure*}[htbp]
\centering
\small
\begin{tcolorbox}[colback=white, colframe=black, title={Example Instruction-Based SFT Sample for Basic Information Extraction},width=0.93\textwidth]
\small
\textbf{Instruction:}
You are a professional resume analysis assistant. Your task is to convert the given resume text into the JSON format specified below. (If both Chinese and English resumes appear, only extract from the Chinese one.)

Extract the following information into a JSON object. If any field does not exist, output an empty string \texttt{""}.

\begin{verbatim}
{
  "basicInfo": {
    "name": "",            # Name, e.g.: Zhang San
    "personalEmail": "",   # Email, e.g.: 610730297@qq.com
    "phoneNumber": "",     # Phone/Mobile number, preserve original format, e.g., "+1 (201) 706 1136"
    "age": "",             # Current age
    "born": "",            # Birth year (and month if available), e.g.: 1996-11
    "gender": "",          # Male/Female. Leave empty if not present.
    "desiredLocation": ["city name", ...], 
                           # Expected job location(s), e.g., ["Beijing", "Shanghai"]
                           # Must be explicitly mentioned. If not, set to [].
    "currentLocation": "", # Current city of residence. Must be explicitly mentioned.
    "placeOfOrigin": ""    # Place of origin. Do not confuse with current location.
  }
}
\end{verbatim}

/no\_think

\medskip
\textbf{Input (Indexed Resume Text):}

[0]: Gu Dabai [1]: Phone: 13987898888 [2]: Email: 123245677@123.com [3]: Job Objective: New Media Operations [4]: Education [5]: 2012.09—2016.06 Hebei University of Science and Technology, School of Business, Major: Information Management and Information Systems, Bachelor's degree [6]: Core Courses: Principles of Management, Macroeconomics, Data Structures, Database Systems, Intro to Software Engineering, CAD [7]: Internet Business Intelligence [8]: Work Experience [9]: 2016.07—Present Rainbow Network Co., New Media Operations Director [10]: Responsible for managing multiple social media accounts (Weibo, WeChat, Toutiao) [11]: Led online/offline brand campaigns and media exposure strategies [12]: Organized celebrity livestreams, reaching 2+ billion views; [13]: Gained ~500K new followers annually per account [14]: 2013.09—2015.06 Haichao Consulting Co., Senior Consultant [15]: Assisted in public relations and brand image development [16]: Participated in project management and business analysis; [17]: Built client profiles and data processing tools [18]: Skills \& Certifications [19]: TEM-8 (Advanced English), experience with Facebook, LinkedIn [20]: National Level 5 JAVA Certificate [21]: Self-Evaluation [22]: Creative, passionate, and curious thinker [23]: Keen on digital and online marketing trends [24]: Highly organized and responsible [25]: Excellent in teamwork and independent execution

\medskip
\textbf{Output:}
\begin{verbatim}
{
  "basicInfo": {
    "name": "Gu Dabai",
    "phoneNumber": "13987898888",
    "personalEmail": "123245677@123.com",
    "age": "",
    "born": "",
    "gender": "",
    "desiredLocation": [],
    "currentLocation": "",
    "placeOfOrigin": ""
  }
}
\end{verbatim}
\end{tcolorbox}
\caption{An Instruction-based SFT Sample for Extracting Basic Personal Information.}
\label{fig:sample_basicinfo}
\end{figure*}

\section{Construction of SynthResume Dataset}
\label{app:dataset}
\begin{figure}[H]
    \centering
\includegraphics[width=\columnwidth]{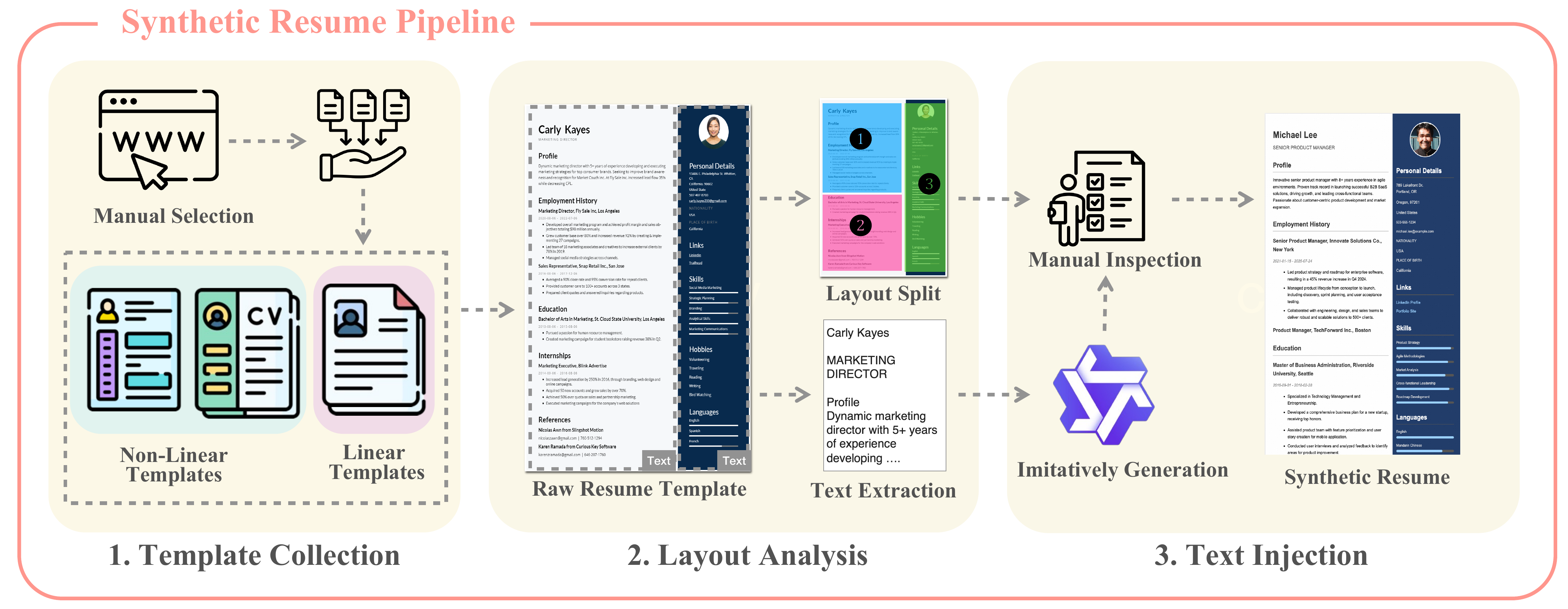}
    \caption{Pipeline of Synthetic Resume Dataset Construction}
    \label{fig:synthResume}
\end{figure}

To facilitate training and evaluation of our models, we construct a large-scale synthetic resume dataset, SynthResume, consisting of 2,994 richly structured samples. This dataset is generated using a semi-automated pipeline as shown in Figure~\ref{fig:synthResume}, designed to simulate realistic resumes with diverse layout and content variations. 

The construction process includes the following key stages:

\sstitle{Step 1: Template Collection.}
We first manually curate a collection of resume templates that reflect a wide variety of real-world styles, including both \textit{linear templates} ( single-column, top-down formats) and \textit{non-linear templates} (complex, multi-column layouts with sidebars). 

\sstitle{Step 2: Layout Analysis and Text Extraction.}  Each resume template is processed through a layout analysis module to identify its major visual blocks (\eg, header, main content column, sidebar) and extract the text within each block. This process resulted in a clean, ordered text representation of the original resume's content, which served as the backbone for synthesizing new resumes.

\sstitle{Step 3: Content Generation and Text Injection.} Each template's extracted content is used as input context to a large language model, with a specific prompt designed to generate new content while preserving the original field structure: ``Given the above resume as a template, generate a new resume by randomly replacing the field values while preserving the structure and semantics.” 
The LLM-generated text undergoes a manual inspection to ensure quality and coherence. Once verified, this new text is injected back into the original visual template. The final output of this stage is a complete, fully-formatted synthetic resume PDF that retains the layout of the source template but contains entirely new, synthetic content.

\sstitle{Finally,} to create high-quality labels for these resumes, we employ a two-pass annotations:
\begin{itemize}[leftmargin=*]
    \item Automated Pre-labeling: Each of the 2,994 synthetic resumes is processed by our full extraction pipeline, using a powerful LLM model Qwen-Max to generate an initial set of indexed text and pre-labeled JSON annotations.
    \item Manual Correction: To ensure label quality, these pre-labels are then subjected to a rigorous manual correction phase by human annotators, correcting any errors made by the initial automated pass.
\end{itemize}
The fully annotated dataset are sorted by the length of the resume text. The longest 2,500 resumes were allocated to the training set, with the remaining 494 forming the test set. This entire process yielded a high-quality, and structurally diverse dataset ideal for training and evaluating layout-aware information extraction models.
\section{Details of Baselines}
\label{app:baseline}
To comprehensively evaluate our proposed framework, we compare it against a wide range of baseline systems, as well as benchmark different LLMs in our pipeline.
\begin{itemize}[leftmargin=*]
    \item \textbf{Bello}~\cite{bello2024}. A widely deployed, commercial resume analysis service in HR automation. It processes resumes across varying formats and layouts, applies bilingual parsing, knowledge graph enhanced field extraction, and document structure understanding at scale. We treat Bello’s Intelligent Parser as a strong industrial baseline in our work.
   % \item \textbf{EzNLP}~\cite{zhu2023deep}: A span-based method that separately models the internal content and the boundary information of a span and then fuses them into a single, highly informative representation for name entity recognition. 
    \item \textbf{PaddleNLP}~\cite{paddlenlp}. An open-source NLP library developed by Baidu based on the PaddlePaddle deep learning framework. It provides pre-trained models and end-to-end pipelines for a wide range of NLP tasks. In our experiments, we adopt its IE workflow for resume field extraction.
    \item \textbf{Claude-4}~\cite{claude4_2024}. The latest large language model from Anthropic, designed for high-performance reasoning and long-context understanding.
    \item \textbf{Gemini-2.5flash}~\cite{gemini2.5_2024}. A lightweight version of Google's Gemini 2.5 model, optimized for fast inference and reduced latency, while maintaining reasonable performance for common LLM tasks.
    \item \textbf{GPT-4o}~\cite{openai2024gpt4technicalreport}. OpenAI’s flagship multimodal model released in 2024, capable of handling text, audio, and image inputs natively.
    \item \textbf{Deepseek-v3}~\cite{deepseek2024}. An open-source, multilingual LLM developed by DeepSeek that achieves competitive performance across various reasoning and language tasks.
    \item \textbf{Qwen-max}~\cite{qwen2024}. A state-of-the-art, multilingual large language model with over 100 billion parameters, released as part of the Qwen model series by Alibaba. It is designed for general-purpose reasoning and excels in multi-step, instruction-following tasks.
    \item \textbf{Qwen3 Series}~\cite{qwen3_2024}. Qwen3 series includes models of various sizes designed for efficiency and controllability. We compare Qwen3-14B, Qwen3-4B, and Qwen3-0.6B to evaluate the performance of our pipeline with different complexity of LLM models.
\end{itemize}
\section{Details of Metrics}
\label{app:metrics}
In our experiments, we adapt three standard evaluation metrics Precision, Recall and F1-score~\cite{deepa2025automated} to resume entity extraction task, and introduce a novel metric Accuracy to specifically evaluate our multi-strategy field matching logic.
Let $E_{gt}$ be the set of ground-truth entity fields, $E_{pred}$ be the set of fields predicted by our model, $E_{align}$ be the set of aligned fields via Hungarian algorithm, and $E_{correct}$ be the set of correct matches through multi-strategy field matching. An entity field $e \in E_{pred}$ is considered correct and included in $E_{correct}$ if and only if it is first successfully aligned with a ground-truth entity via the Hungarian algorithm (\ie, in $E_{align}$) and subsequently passes our multi-strategy field matching criteria.

Based on these definitions, we compute the following metrics:
\begin{itemize}[leftmargin=*]
\item \textbf{Precision} measures the proportion of correctly extracted entity fields among all fields predicted by the model.
\begin{equation}
    \text{Precision} = \frac{|E_{correct}|}{|E_{pred}|}
\end{equation}
\item \textbf{Recall} quantifies the proportion of ground-truth entity fields that are correctly extracted by the model.
\begin{equation}
    \text{Recall} = \frac{|E_{correct}|}{|E_{gt}|}
\end{equation}
\item \textbf{F1-Score} is the harmonic mean of Precision and Recall, providing a single, balanced measure of overall performance.
\begin{equation}
    \text{F1-Score} = 2 \times \frac{\text{Precision} \times \text{Recall}}{\text{Precision} + \text{Recall}}
\end{equation}
\item \textbf{Accuracy} measures the fraction of correctly extracted entity fields among all aligned fields. This metric helps us distinguish errors caused by the alignment algorithm from those caused by our field-matching rules.
\begin{equation}
    \text{Accuracy} = \frac{|E_{correct}|}{|E_{align}|} 
\end{equation}
\end{itemize}

\section{Finer-Grained Comparison of Different Models}
\label{app:finegrained}
In addition to the overall performance averaged across all resume fields, we also conduct finer-grained accuracy comparisons across different field groups in SynthResume and RealResume datasets. The complete results are illustrated in Table~\ref{tab:synthResume_results} and Table~\ref{tab:realresume_results} respectively.

\begin{table*}[h]
\centering
\caption{Fine-grained Performance Comparison of Different model on SynthResume Dataset.}
\label{tab:synthResume_results}
\resizebox{0.9\textwidth}{!}{
\begin{tabular}{l rrrr rrrr rrrr}
\toprule
& \multicolumn{4}{c}{\textbf{Period}} & \multicolumn{4}{c}{\textbf{Named Entity}} & \multicolumn{4}{c}{\textbf{Long Text}} \\
\cmidrule(lr){2-5} \cmidrule(lr){6-9} \cmidrule(lr){10-13}
\textbf{Model} & \textbf{Acc.} & \textbf{Prec.} & \textbf{Recall} & \textbf{F1} & \textbf{Acc.} & \textbf{Prec.} & \textbf{Recall} & \textbf{F1} & \textbf{Acc.} & \textbf{Prec.} & \textbf{Recall} & \textbf{F1}\\
\midrule
\multicolumn{9}{l}{\textit{Non-LLM Baselines}} \\
\quad Bello & 0.852 & 0.922 & 0.852 & 0.885 & 0.841 & 0.781 & 0.720 & 0.749& 0.469 & 0.246 & 0.273 & 0.259\\
\quad PaddleNLP & 0.433 & 0.723 & 0.407 & 0.511 & 0.818 & 0.727 & 0.679 & 0.699 & 0.272 & -- & -- & -- \\
\midrule
\multicolumn{9}{l}{\textit{OCR + LLM}}\\
\quad Claude-4 & 0.977  & 0.984 &	0.975 &	0.980 & 0.939 &	0.955 &	0.951 &	0.951 & 0.731   & 	0.634 &	0.741 &	0.684 \\
\midrule
\multicolumn{9}{l}{\textit{Our Pipeline (LLM)}} \\
\quad Claude-4 & 0.960 & 0.983 & 0.963 & 0.973 & 0.963 & 0.964 & 0.949 & 0.956 & 0.825 & 0.786 & 0.816 & 0.801\\
\quad Gemini2.5-flash & 0.963 & 0.983 & 0.967 & 0.975 & 0.959 & 0.972 & 0.944 & 0.957 & 0.835 & 0.823 & 0.842 & 0.831\\
\quad GPT-4o & 0.960 & 0.987 & 0.961 & 0.974 & 0.966 & 0.968 & 0.955 & 0.961 & 0.840 & 0.819 & 0.840 & 0.829\\
\quad Deepseek-v3 & 0.961 & 0.978 & 0.965 & 0.971 & 0.968 & 0.976 & 0.951 & 0.962 & 0.829 & 0.834 & 0.794 & 0.813\\
\quad Qwen-max & 0.942 & 0.948 & 0.943 & 0.945 & 0.979 & 0.969 & 0.970 & 0.970 & 0.824 & 0.798 & 0.826 & 0.811\\
\quad Qwen3-14B & 0.888 & 0.892 & 0.913 & 0.902 & 0.961 & 0.955 & 0.954 & 0.954 & 0.698 & 0.664 & 0.665 & 0.664\\
\quad Qwen3-4B & 0.893 & 0.872 & 0.907 & 0.889 & 0.955 & 0.949 & 0.956 & 0.953 & 0.513 & 0.503 & 0.554 & 0.527 \\
\quad Qwen3-0.6B & 0.619 & 0.682 & 0.660 & 0.668 & 0.637 & 0.749 & 0.734 & 0.691 & 0.256 & 0.162 & 0.226 & 0.184\\
\midrule
\multicolumn{9}{l}{\textit{Our Pipeline (SFT)}} \\
\quad Qwen3-0.6B-sft & 0.897 & 0.908 & 0.907 & 0.907 & 0.951 & 0.941 & 0.936 & 0.938 & 0.858 & 0.760 & 0.777 & 0.767\\
\bottomrule
\end{tabular}
}
\end{table*}

\begin{table*}[h]
\centering
\caption{Fine-grained Performance comparison of Different model on RealResume Dataset.}
\label{tab:realresume_results}
\resizebox{0.9\textwidth}{!}{
\begin{tabular}{l rrrr rrrr rrrr}
\toprule
& \multicolumn{4}{c}{\textbf{Period}} & \multicolumn{4}{c}{\textbf{Named Entity}} & \multicolumn{4}{c}{\textbf{Long Text}} \\
\cmidrule(lr){2-5} \cmidrule(lr){6-9} \cmidrule(lr){10-13}
\textbf{Model} & \textbf{Acc.} & \textbf{Prec.} & \textbf{Recall} & \textbf{F1} & \textbf{Acc.} & \textbf{Prec.} & \textbf{Recall} & \textbf{F1} & \textbf{Acc.} & \textbf{Prec.} & \textbf{Recall} & \textbf{F1}\\
\midrule
\multicolumn{9}{l}{\textit{Non-LLM Baselines}} \\
\quad Bello & 0.921 & 0.968 & 0.813 & 0.879 & 0.885 & 0.801 & 0.740 & 0.769 & 0.540 & 0.553 & 0.459 & 0.500 \\
\quad PaddleNLP & 0.387 & 0.587 & 0.381 & 0.451& 0.722 & 0.653 & 0.597 & 0.622   & -- & -- & -- & -- \\
\midrule 
\multicolumn{9}{l}{\textit{OCR + LLM}} \\
\quad Claude-4 &  0.979 & 0.987 &	0.984 	&0.986 & 0.937 & 0.958 &	0.960 &	0.959 & 0.582 & 0.512 &	0.598 &	0.548 \\
\midrule
\multicolumn{9}{l}{\textit{Our Pipeline (LLM)}} \\
\quad Claude-4 & 0.963 & 0.985 & 0.960 & 0.972& 0.964 & 0.924 & 0.980 & 0.949 & 0.869 & 0.819 & 0.899 & 0.854 \\
\quad Gemini2.5-flash & 0.970 & 0.984 & 0.973 & 0.978& 0.945 & 0.898 & 0.974 & 0.931  & 0.888 & 0.851 & 0.880 & 0.865 \\
\quad GPT-4o & 0.963 & 0.982 & 0.962 & 0.972 & 0.950 & 0.914 & 0.973 & 0.941& 0.880 & 0.848 & 0.899 & 0.870 \\
\quad Deepsek-v3 & 0.960 & 0.987 & 0.957 & 0.971& 0.939 & 0.903 & 0.940 & 0.918 & 0.867 & 0.842 & 0.865 & 0.852 \\
\quad Qwen-max & 0.971 & 0.989 & 0.968 & 0.978 & 0.936 & 0.895 & 0.941 & 0.915  & 0.827 & 0.810 & 0.832 & 0.820 \\
\quad Qwen3-14B & 0.963 & 0.982 & 0.963 & 0.972 & 0.939 & 0.871 & 0.935 & 0.899 & 0.724 & 0.699 & 0.717 & 0.707 \\
\quad Qwen3-4B & 0.901 & 0.902 & 0.963 & 0.930 & 0.921 & 0.849 & 0.936 & 0.886  & 0.579 & 0.533 & 0.612 & 0.567 \\
\quad Qwen3-0.6B & 0.680 & 0.750 & 0.724 & 0.734 & 0.647 & 0.683 & 0.697 & 0.671 & 0.120 & 0.126 & 0.182 & 0.136 \\
\midrule
\multicolumn{9}{l}{\textit{Our Pipeline (SFT)}} \\
\quad Qwen3-0.6B-sft & 0.956 & 0.976 & 0.951 & 0.963 & 0.953 & 0.909 & 0.962 & 0.932  & 0.866 & 0.807 & 0.874 & 0.838 \\
\bottomrule
\end{tabular}
}
\end{table*}

\end{document}